\documentclass[letterpaper]{article} 
\usepackage{aaai23}  
\usepackage{times}  
\usepackage{helvet}  
\usepackage{courier}  
\usepackage[hyphens]{url}  
\usepackage{graphicx} 
\usepackage{natbib}  
\usepackage{caption} 
\usepackage{graphicx}
\usepackage{subfigure}
\usepackage{amssymb}
\usepackage{amsmath}
\usepackage{multirow}
\usepackage{color}
\usepackage{todonotes}
\usepackage{bbding}

\usepackage{algorithm}
\usepackage{algorithmic}

\usepackage{newfloat}
\usepackage{listings}
\DeclareCaptionStyle{ruled}{labelfont=normalfont,labelsep=colon,strut=off} 
\floatstyle{ruled}
\newfloat{listing}{tb}{lst}{}
\floatname{listing}{Listing}
 \nocopyright 

\title{Robust Domain Adaptation for Machine Reading Comprehension}
\author{
	Liang Jiang\equalcontrib\textsuperscript{\rm 1},
	Zhenyu Huang\equalcontrib\textsuperscript{\rm 2},
	Jia Liu\textsuperscript{\rm 1},
	Zujie Wen\textsuperscript{\rm 1},
	Xi Peng\textsuperscript{\rm 2}\thanks{Corresponding author.}
}
\affiliations{
    \textsuperscript{\rm 1}Ant Group\\
	\textsuperscript{\rm 2}College of Computer Science, Sichuan University\\
	\{tianxuan.jl, jianiu.lj, zujie.wzj\}@antgroup.com, \{zyhuang.gm, pengx.gm\}@gmail.com\\

}

\usepackage{bibentry}

\begin{document}

\maketitle

\begin{abstract}
Most domain adaptation methods for machine reading comprehension (MRC) use a pre-trained question-answer (QA) construction model to generate pseudo QA pairs for MRC transfer. Such a process will inevitably introduce mismatched pairs (i.e., noisy correspondence) due to i) the unavailable QA pairs in target documents, and ii) the domain shift during applying the QA construction model to the target domain. Undoubtedly, the noisy correspondence will degenerate the performance of MRC, which however is neglected by existing works. To solve such an untouched problem, we propose to construct QA pairs by additionally using the dialogue related to the documents, as well as a new domain adaptation method for MRC. Specifically, we propose Robust Domain Adaptation for Machine Reading Comprehension (RMRC) method which consists of an answer extractor (AE), a question selector (QS), and an MRC model. Specifically, RMRC filters out the irrelevant answers by estimating the correlation to the document via the AE, and extracts the questions by fusing the candidate questions in multiple rounds of dialogue chats via the QS. With the extracted QA pairs, MRC is fine-tuned and provides the feedback to optimize the QS through a novel reinforced self-training method. Thanks to the optimization of the QS, our method will greatly alleviate the noisy correspondence problem caused by the domain shift. To the best of our knowledge, this could be the first study to reveal the influence of noisy correspondence in domain adaptation MRC models and show a feasible way to achieve robustness to mismatched pairs. Extensive experiments on three datasets demonstrate the effectiveness of our method.
\end{abstract}

\section{Introduction}
In recent, a number of domain adaptation (DA) methods~\cite{cao2020unsupervised,wang2019adversarial,lewis2019unsupervised} for machine reading comprehension (MRC) have been proposed, which usually pre-train an MRC model in a high-resource domain and then transfer it to a low-resource domain. 
Specifically, most existing methods~\cite{cao2020unsupervised,wang2019adversarial,lewis2019unsupervised} consist of two steps. First, they construct some pseudo QA pairs by using a pre-trained QA construction model from the available documents in the target domain. Then they fine-tune the pre-trained MRC model by using the constructed pairs. 

\begin{figure*}[!hbt]
\centering
	\subfigure[Noisy Labels vs. Noisy Correspondence] {
	\label{fig:msn}
	\includegraphics[width=0.46\textwidth]{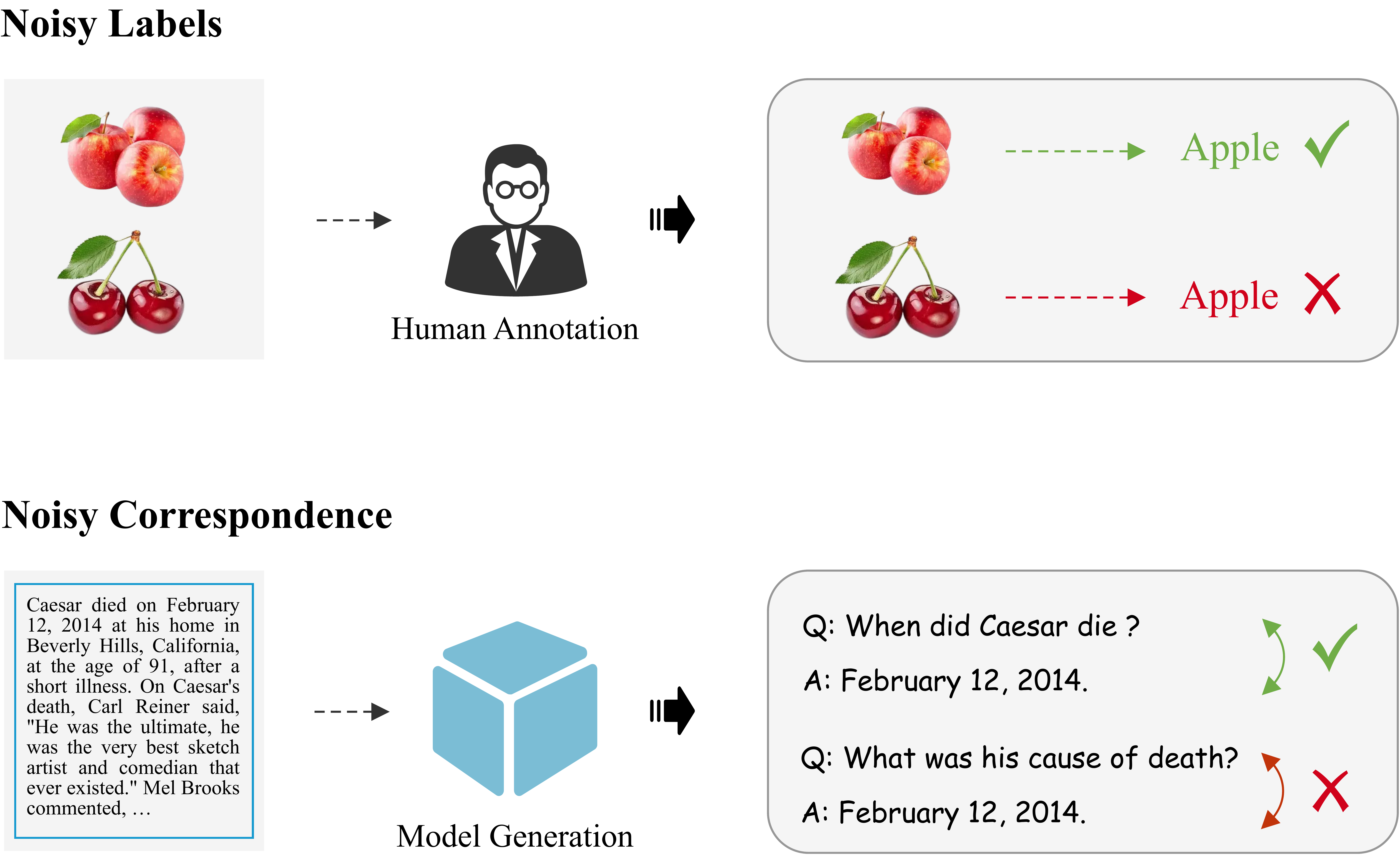}
	}
	\hspace{0.6cm}
	\subfigure[Existing Works vs. Our Work] {
	\label{fig:motivation}
	\includegraphics[width=0.43\textwidth]{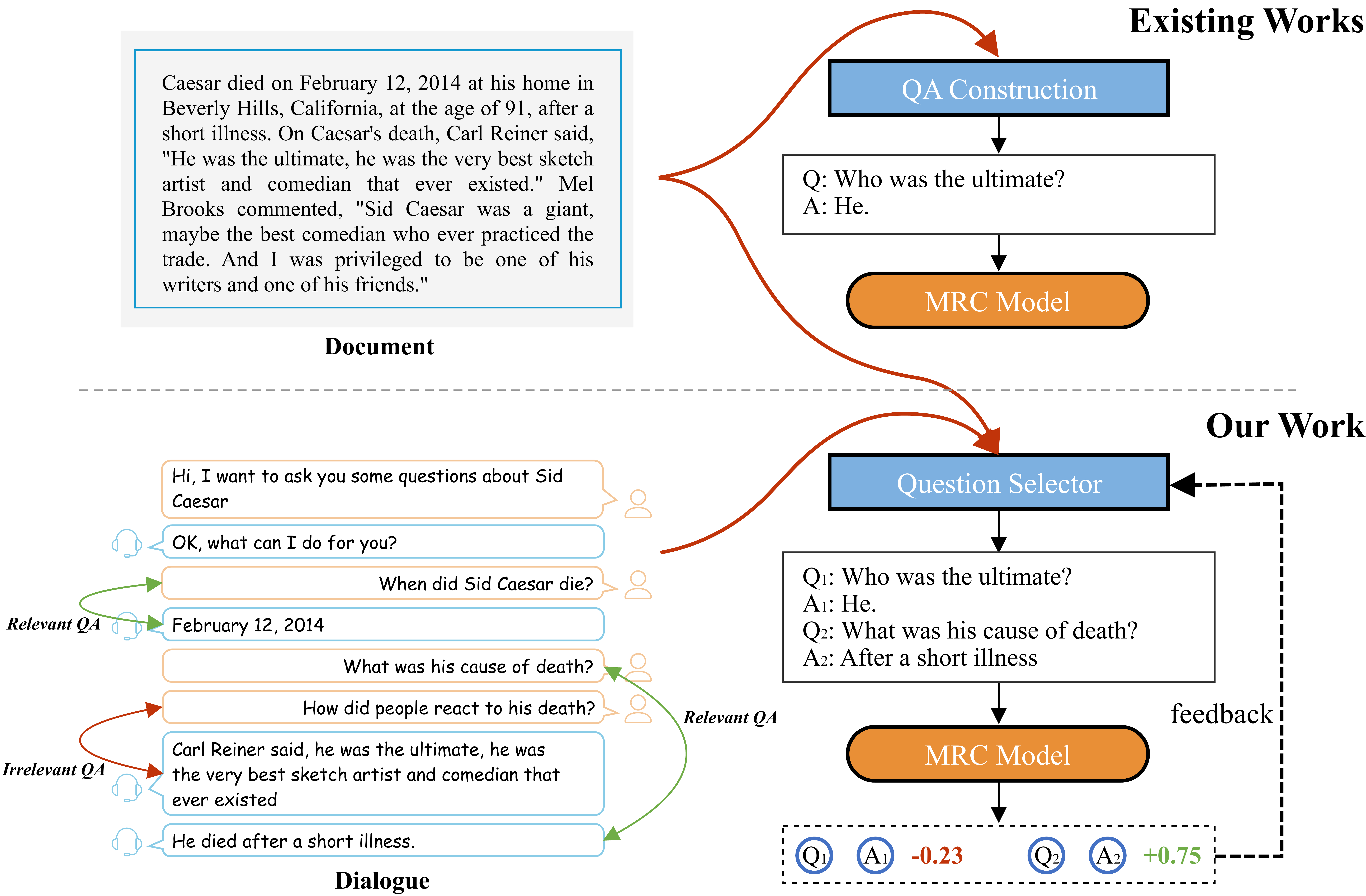}
	}
\vskip -0.2cm
  \caption{(a) Noisy Labels vs. Noisy Correspondence. The noisy labels refer to the errors in the category annotation of data samples caused by human annotation. The noisy correspondence here refers to the misalignment between two data points generated by the model itself. (b) An example of the document along with dialogue, and the difference between our method and the existing ones. The dialogue is the conversations between the questioner and the answerer (the customer and customer service here) about the document. In other words, the questioner raises a question about the document, and the answerer answers it by referring to the document. Hence, the dialogues have natural QA-form conversations that is helpful in QA pair construction. }
\vskip -0.4cm
\end{figure*}

Although these methods have achieved promising results, almost all of them ignore the mismatched QA pairs, i.e., the irrelevant QA pairs which are wrongly treated as positive. In the scenario of MRC, such a so-called noisy correspondence (NC) issue~\cite{huang2021learning} is caused by the following reasons. First, the domain adaptation methods for MRC often construct pseudo QA pairs by using the documents in the target domain which does not contain natural questions. As a result, the generated questions will be probably irrelevant to the answers. Second, the QA construction model is pre-trained in the source domain and then directly applied to the target domain without fine-tuning. In consequence, such a domain shift issue will lead to noisy correspondence. A toy example of NC is shown in Fig.~\ref{fig:msn}, and more real-world samples generated by the existing works refer to Fig.~\ref{fig:case_study}. Notably, the NC problem is remarkably different from the well-studied noisy labels. To be specific, noisy labels generally refer to the category-level annotation errors of a given data point, whereas here NC refers to the mismatched relationship between two data points. Undoubtedly, NC will degenerate the performance of the MRC model, which is however neglected so far as we know.

To sum up, domain adaptation methods for MRC will face the NC challenge caused by i) the unavailable QA pairs in the target documents, and ii) domain shift during applying the QA construction model to the target domain. In this paper, we propose to construct more credible QA pairs by additionally using document-associated dialogues to fill the vacancy of the natural question information in the document. In addition, our method will fine-tune the QA construction model in the target domain with the help of the MRC model feedback. For clarity, we summarize the major differences between our MRC method and the existing ones in Fig.~\ref{fig:motivation} by taking the customer service as a showcase without loss of generality. As shown, in real-world applications, the customer service usually answers customers' questions by referring to the documents, forming an associated dialogue. Existing works only use the sole document for QA construction and do not further fine-tune the QA construction model in the target domain. In contrast, our work leverages both documents and the associated dialogues for QA construction. Dialogues are the conversation corpus between the questioners and answerers, which naturally preserves QA-similar chats and thus are more credible for QA construction. Moreover, our method will use the feedback of the MRC model on constructed QA pairs to optimize the QA construction model, thus alleviating the domain shift issue and improving the QA quality.

In practice, however, difficulties arise when an attempt is made to apply the above approaches. Specifically, although dialogues are more credible than documents for the QA construction, they still contain a huge number of irrelevant and discontinuous conversations. In other words, the above QA construction method could partially alleviate but still cannot solve the NC problem. As shown in Fig.~\ref{fig:motivation}, the irrelevant conversation is about the irrelevant conversations to the document, e.g., there are some greetings or other chats which are irrelevant w.r.t the document. The discontinuous conversation is that the question and answer are not exactly aligned in a single round of chat due to the complexity in interaction, e.g., the customer may raise a new question before receiving the answer for the last ones. Besides the challenges rooted in the dialogue data, another difficulty is how to fine-tune the non-differentiable QA construction model in the target domain to alleviate the domain shift issue. In brief, the existing works often generate QA pairs by resorting to the discrete $\mathop{\arg \max}$ sampling operator which hinders the optimization of the QA construction model. 

To overcome the above challenges in data quality and model optimization, we propose a novel domain adaptation method for MRC, dubbed Robust Domain Adaptation for Machine Reading Comprehension (RMRC) which consists of an answer extractor (AE), a question selector (QS), and an MRC model. In brief, RMRC leverages both documents and the associated dialogue for MRC transfer by i) constructing QA pairs via the AE and the QS, and ii) alternative training the MRC model and the QS. 
In the first stage, for a given document, RMRC extracts the candidate answers from dialogues and filters out unrelated ones in terms of the estimated relevance using the pre-trained AE, thus alleviating the NC issue caused by irrelevant chats. After that, for the extracted answer, RMRC seeks the most related questions that appeared in multiple rounds of chats via the pre-trained QS, thus tackling the NC problem caused by discontinuous conversations. 
In the second stage, RMRC optimizes the MRC model and the QS in an alternative fashion, which will favor MRC transfer. In detail, the domain shift could be alleviated and accordingly the NC problem is tackled by optimizing the QS using the feedback of the MRC model.  
Note that, as the QS is non-differentiable and cannot be directly optimized by back-propagation, we propose a novel reinforced self-training optimization method by recasting the model evaluation on the constructed QA pairs as a training reward.

The main contributions and novelty of this paper could be summarized as below. 
\begin{itemize}
    \item This work could be the first successful attempt to study the NC problem that is common but ignored in existing domain adaptation methods for MRC. 
    \item To solve the NC problem, we propose to leverage both the document and associated dialogue for MRC model training. To the best of our knowledge, this is the first method on study how to leverage the dialogue in the domain adaptation for MRC.
    \item To implement robust domain adaptation method for MRC, RMRC consists of the AE, QS, and MRC model. Thanks to the reinforced self-training optimization method, the QS could be fine-tuned with the MRC model feedback, thus further alleviating the influence of NC. 
\end{itemize}

\section{Related Work}
In this section, we will briefly introduce some recent developments in machine reading comprehension, the domain adaptation for MRC, and noisy label learning.

\subsection{Machine Reading Comprehension}
Machine reading comprehension aims to read the documents and answer questions about a document. Thanks to the collected benchmark datasets like SQuAD~\cite{rajpurkar2016squad}, CoQA~\cite{reddy2019coqa}, and QuAC~\cite{choi2018quac}, MRC has made great progress in recent years, and even surpasses the human level~\cite{seo2016bidirectional,yu2018qanet,devlin2018bert,zhang2020retrospective,jiang2018stackreader,gao2019product}. 
\cite{seo2016bidirectional} proposed BiDAF which leverages RNN and bi-directional attention mechanism between question and document to achieve promising performance in machine reading comprehension. QANet~\cite{yu2018qanet} used CNN rather than RNN to better capture local information in question and documents. 
BERT~\cite{devlin2018bert} proposes to use a large amount of unsupervised corpus for pre-training and successfully improve performance on many downstream NLP tasks including MRC. 


\subsection{Domain Adaptation}
Domain adaptation (DA) is a well-developed technique that aims to transfer knowledge in the presence of the domain gap, i.e. making a model pre-trained in the source domain generalizable to the target domain~\cite{zhang2018collaborative,zhang2021meta,ding2018graph,hu2018duplex,kan2015bi,li2013learning}.
The existing domain adaptation methods for MRC could be roughly grouped into two categories: 

1) \textbf{Model generalization.} It aims to improve the generalization capability of the model trained on the source domain to the target domain~\cite{li2019d,su2019generalizing,baradaran2021ensemble}. 
For example, \cite{su2019generalizing} propose to use a pre-trained model trained on multiple MRC datasets simultaneously to improve the generalization ability on the new domains.
\cite{baradaran2021ensemble} propose to ensemble the models individually trained on different datasets to improve the generalization ability.
However, this kind of methods do not leverage the available information on the target domain, thus obtaining less satisfactory performance. 

2) \textbf{QA generation.} It often utilizes the target-domain documents to construct QA pairs to fine-tune the MRC models~\cite{du2017learning,wang2019adversarial, lewis2019unsupervised,cao2020unsupervised}. 
For example, \cite{du2017learning} propose to generate questions on the target domain using a Seq2seq question generator that trained on the source domain, and then fine-tune the MRC model with the pseudo QA pairs.
\cite{cao2020unsupervised} propose to add a self-training step to filter out low quality QA pairs using the estimated scores of each constructed pairs through the MRC model.
Though these methods have made great progress, they still suffer from the noisy correspondence problem as discussed in the introduction, thus leading to sub-optimal performance in real-world applications. 
Unlike these works, to address the NC problem, this paper proposes to construct more credible QA pairs by additionally using the dialogue, and optimize the MRC model and the question selector in an alternative fashion to alleviate the domain shift of the QS to the target domain.


\subsection{Learning with Noisy Labels}
To alleviate even eliminate the influence of noisy labels, many methods have been proposed to achieve robust classification results~\citep{arazo2019unsupervised,song2020learning,liu2015classification,xia2020robust,bai2021understanding,luo2021bi}. Currently, the existing works often resort to sample selection for achieving noise robustness. 
To be specific, sample selection methods seek to select the clean samples from the noisy dataset for training. For example, \cite{arazo2019unsupervised} proposes to select the samples with small loss as the clean samples. 
Moreover, to further enhance the clean sample selection capacity, Co-teaching methods~\citep{han2018co,yu2019does} leverage two individual trained networks to filter out noises in an alternative manner. 
In recent, PES~\citep{bai2021understanding} proposes a progressive early stopping strategy in the semi-supervised learning framework by treating the clean and noisy samples as the labeled and unlabeled data. 
In addition, some very recent works~\cite{yang2021partially,huang2021learning} study the paradigm of noisy correspondence like this paper. Although the works share some similarity with this work, they are significantly different in motivation, application, and method.

\begin{figure*}[t]
\centering
\includegraphics[width=0.92\textwidth]{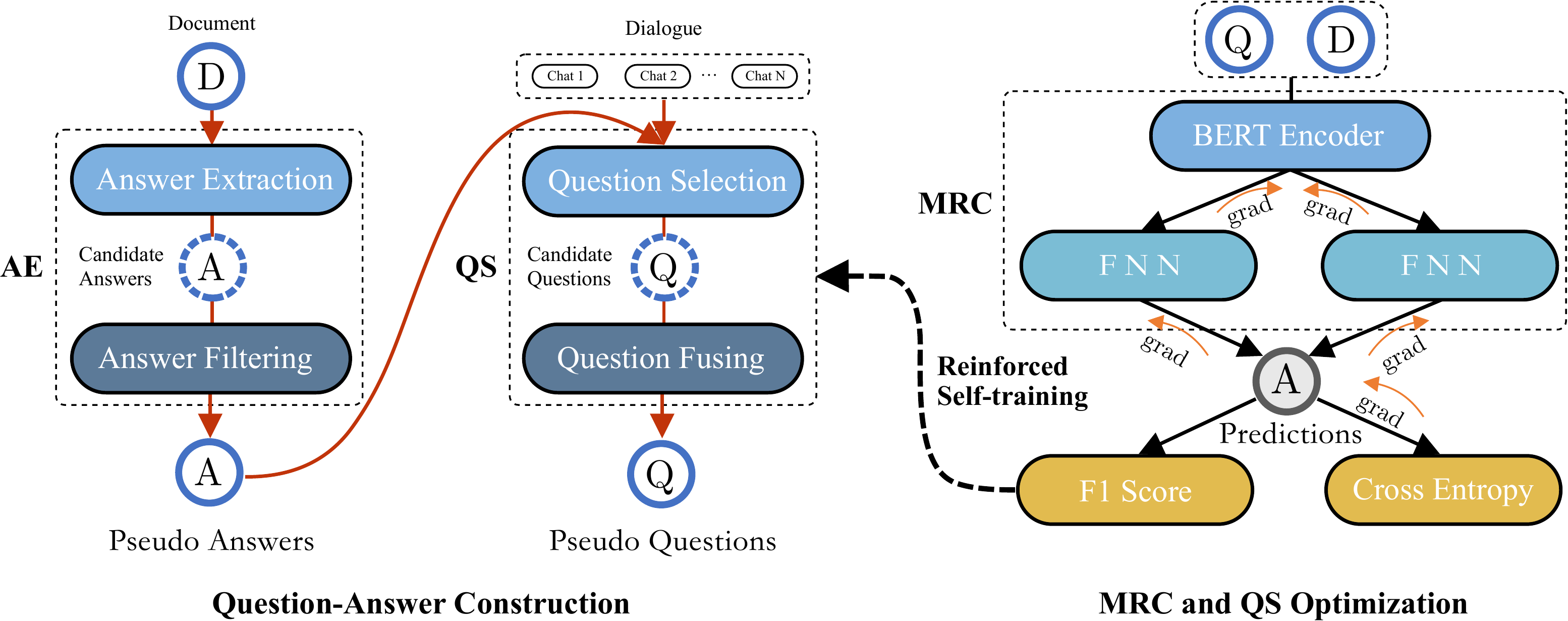}
 \vskip -0.1cm
\caption{Overview of our method. RMRC consists of an answer extractor (AE), a question selector (AS), and an MRC model. In the first stage, AE and QS construct the QA pairs in the target domain by leveraging both the documents and dialogues. In the second stage, RMRC optimizes the MRC model and QS in an alternative fashion. }
\label{RMRC_pipeline}
 \vskip -0.4cm
\end{figure*}

\section{Method}
In this section, we elaborate on the proposed Robust Domain Adaptation for Machine Reading Comprehension (RMRC). 

\subsection{Problem Definition}
\label{sec:problem_definition}
For a given document and a related question, MRC aims to find a text span from the document as the corresponding answer. To overcome the data scarcity issue in MRC, some domain adaptation methods are proposed to transfer an MRC model pre-trained in the source domain $\mathcal{S}$ to the target domain $\mathcal{T}$.
Different from existing domain adaptation methods for MRC, we utilize the documents along with the associated dialogues in $\mathcal{T}$ to construct QA pairs for model fine-tuning. 
Formally, given a pre-trained MRC model $M$ and $B$ documents $\mathcal{D} = \{D_1, D_2, \cdots, D_B\}$ in the target domain, where each document $D_b$ is associated with a dialogue set generated by the questioners and answerers, we aim to transfer $M$ from $\mathcal{S}$ to $\mathcal{T}$. For each dialogue $ C =\{C^{r_1}_1, C^{r_2}_2, \cdots, C^{r_{N_c}}_{N_c} | r_i \in (a, q)\}$, it contains $N_c$ chats, where $r_i$ denotes the speaker of the $i$-th chat, $a$ and $q$ indicate the role of answerer and the questioner, respectively.

For clarity, we first provide an overview of the proposed RMRC and then introduce its components one by one. As shown in Fig.~\ref{RMRC_pipeline}, RMRC consists of an answer extractor (AE), a question selector (QS), and an MRC model which are applied to the following two stages: 1) \textbf{QA construction with AE and QS}. For each given document and the associated dialogues, the AE first splits the document into a set of candidate answers. Then, the AE outputs the pseudo answers by filtering out the irrelevant answers in terms of the relevance to the corresponding chat. After that, for each given pseudo answer and the related chat, QS selects multiple chats located before the answer-related chat in the dialogue as the candidate questions. Finally, the most related candidate questions are concatenated as pseudo questions. 2) \textbf{Alternative training of the MRC model and QS}. For a given pseudo question and document, RMRC uses the MRC model to obtain the corresponding answer. After that, the obtained answer is used to optimize the MRC model via a cross-entropy loss and the QS via our novel reinforce self-training optimizer.


\subsection{Answer Extractor}
\label{sec:answer_extractor}

The AE is designed to extract answers by finding the most similar text span from the document in terms of the answerer's response contained in the dialogue. Formally, with the document set $\mathcal{D}$ and associated dialogue $C$, AE extracts the corresponding answer $\{\mathbf{a}_i\}_{i=1}^{N} \in \mathcal{D}$ by filtering out the unrelated answers. 
Specifically, we first split all documents in $\mathcal{D}$ into $\mathrm{n\text{-}gram}$ tokens $G$ via
\begin{equation}
\label{eq:ngram}
\begin{aligned}
    G &= \{G_1 \cup \cdots \cup G_B\} \\
    G_b &= \{G_b^k\}_{k=1}^K, G_b^k = \mathrm{n\text{-}gram}(D_b, k),
\end{aligned}
\end{equation}
where $\mathrm{n\text{-}gram}$ is the operator to extract the candidate answers from $\mathcal{D}$, $K$ denotes the maximum token number of the candidate answers, $G_b^k$ denotes the $k$-gram token set from document $G_b$. In other words, we will extract  $\mathrm{1\text{-}gram}, \mathrm{2\text{-}gram}, \cdots, \mathrm{K\text{-}gram}$ tokens for answer extraction.
With the candidate answers, a text matching function is developed to find the best matched answer $\mathbf{g}_i \in G$ to $C^{a}_t$ which denoted as $\mathbf{a}_t$, i.e., 
\begin{equation}
\label{eq:matching}
    \mathbf{a}_t, D_t^* = \mathop{\arg \max}_{\mathbf{g}_i \in G_b, G_b\in G} S(C^{a}_t, \textbf{g}_i),
\end{equation}
where $\mathbf{g}_i$ is the extracted n-gram token in $G$, $(\mathbf{a}_t, D_t^*)$ are the best matched answer and the  corresponding document to $C^{a}_t$.
In the equation, we compute the cosine similarity $S(C^{a}_t, \mathbf{g}_i)$ between the extracted features of $C^{a}_t$ and $\mathbf{g}_i$ from a pre-trained BERT~\cite{devlin2018bert}. 
However in the real-world dialogues, the conversation contains not only the document-related answers but also some unrelated answers like greetings as shown in Fig.~\ref{fig:motivation}. Hence, to filter out these unrelated answers, we select the answer $\mathbf{a}_t$ whose matching score is larger than a given threshold $\gamma$. Formally, 
\begin{equation}
\label{eq:a_pick}
    A = \{(\mathbf{a}_t, D_t^*, C_t^a) | \gamma \leq S(C^{a}_t, \mathbf{a}_t), \forall C^{a}_t \in C\}.
\end{equation}



\subsection{Question Selector}
\label{sec:QA_selector}

For a given answer $\mathbf{a}_t$, we pass it and its associated document $D_t^*$ along with the corresponding chat $C^a_t$ through the QS to find the corresponding question $\mathbf{q}_t$ from the dialogue. 
In detail, we first obtain the candidate questions by selecting multiple chats located before $C^{a}_t$ in the dialogue, i.e., 
\begin{equation}
\label{eq:q_candidate}
    \tilde{Q} = \{C^q_{t'}\vert C^q_{t'} \in \mathcal{N}^{\tau}(C^{a}_t)\},
\end{equation}
where $\mathcal{N}^{\tau}(C^{a}_t)$ denotes $\tau$ closest chats located before $C^{a}_t$ from questioner, $\tau$ is the maximal question selection range which is fixed to $16$ throughout our experiments. Such a question selection strategy is based on the observation that the corresponding questions only exist before the answers and are more irrelevant as they are further away from the answer.
For each $C^{q}_{t'} \in \tilde{Q}$, we then compute the relevance score $R(C^{a}_t, C^{q}_{t'})$ between the answer $C^{a}_t$ and the question $C^{q}_{t'}$ by
\begin{equation}
\begin{aligned}
    R(C^{a}_t, C^{q}_{t'}) & =\mathrm{Sigmoid}(h^{qa}_{t'}), \quad
    h^{qa}_{t'} & = \mathrm{Encoder}_Q([C^{a}_t; C^{q}_{t'}])
\label{eq:qa_relevance}
\end{aligned}
\end{equation}
where $\mathrm{Encoder}_Q$ transforms $[C^{a}_t; C^{q}_{t'}]$ into a hidden vector for relevance prediction, $[;]$ denotes the concatenation operator. Note that, the relevance score $R(C^{a}_t, C^{q}_{t'})$ could be regarded as the conditional probability of $C^{q}_{t'}$ to be the corresponding question of the given answer $C^{a}_t$, i.e., $P(C^{q}_{t'}|C^{a}_t) \Leftrightarrow R(C^{a}_t, C^{q}_{t'})$.
To further alleviate the aforementioned discontinuous conversation issue in the real world (e.g., the questioner may raise a new question before the last one is answered.), we concatenate the most related questions in terms of the relevance score w.r.t. $C^{a}_t$, i.e., 
\begin{equation}
\label{eq:q_pick}
\begin{aligned}
\mathbf{q}_t = & \mathrm{Concat}(\{ C^{q}_{t'} | C^{q}_{t'} \in \mathcal{N}^{\kappa}(C_t^a)\}),
\end{aligned}
\end{equation}
where $\mathcal{N}^{\kappa}(C_i^q)$ denotes $\kappa$-nearest neighbors of $C^a_t$ in $\tilde{Q}$ in terms of the relevance score. Accordingly, the probability of $\mathbf{q}_t$ to be the corresponding question of $C_t^a$ is formulated as,
\begin{equation}
P(\mathbf{q}_t|C_t^a)=\prod_{C_{t'}^q \in \mathcal{N}^{\kappa}(C_t^a)}P(C_{t'}^q|C_t^a).
\end{equation}

\subsection{MRC Model Training}
\label{sec:training}
With the constructed pseudo QA pairs and the corresponding document $\{(\mathbf{a}_i, \mathbf{q}_i, D_i)\}_{i=1}^{N}$, we fine-tune the pre-trained MRC model to improve its generalization to the target domain. In detail, we first embed the question and associated document into a hidden space with a BERT encoder, i.e.,
\begin{equation}
\mathbf{v}_i = \mathrm{Encoder_{MRC}}([D_i; \mathbf{q}_i]).
\label{mrc_encode}
\end{equation}

Then, $\mathbf{v}_i$ is utilized to predict the positions of the corresponding answer in the document. Specifically,
\begin{equation}
\begin{aligned}
\label{mrc_prob}
P^{s}_i & = \mathrm{Softmax}(\mathrm{FFN}(\mathbf{W}^{s}; \mathbf{v}_i)),\\
P^{e}_i & = \mathrm{Softmax}(\mathrm{FFN}(\mathbf{W}^{e}; \mathbf{v}_i)),
\end{aligned}
\end{equation}
where $P_i^{s}$ and $P_i^{e}$ denote the probability of each token being the start and end positions of the answer in $D_i$, respectively. $\mathrm{FFN}(\mathbf{W}^{s}; \mathbf{v})$ and $\mathrm{FFN}(\mathbf{W}^{e}; \mathbf{v})$ are two one-layer feed-forward networks with parameters $\mathbf{W}^{s}$ and $\mathbf{W}^{e}$, respectively.
Finally, we use the cross entropy between the predicted probability $(P_i^{s}, P_i^{e})$ and the ground truth as the training loss for our MRC model, i.e.,
\begin{equation}
\label{eq:loss_mrc}
    \mathcal{L}_{MRC} = \frac{1}{N}\sum_i^{N} \mathrm{CE}(P_i^{s}, Y^{s}_i) + \mathrm{CE}(P_i^{e}, Y^{e}_i),
\end{equation}
where $Y^{s}_i$ and $Y^{e}_i$ are two one-hot vectors which denote the start and end positions of $\mathbf{a}_i$ in $D_i$.

\subsection{Reinforced Self-training for QS}
\label{sec:training_for_qa}
As the QS is pre-trained on the source domain, the noisy correspondence will be inevitably introduced when the QS is applied to the target domain for constructing QA pairs. 
To address this domain shift issue, we propose to optimize the QS in the target domain by recasting the model evaluation on the constructed pseudo QA pairs as a training reward. 
Specifically, for a given pair $(\mathbf{a}_i, \mathbf{q}_i, D_i)$, we first obtain the predicted answer by
\begin{equation}
\begin{aligned}
\tilde{\mathbf{a}}_i &= [t_{\tilde{s}}:t_{\tilde{e}}], t \in D, \\
\tilde{s} &= \mathop{\arg \max} P_i^{s}, \\
\tilde{e} &= \mathop{\arg \max} P_i^{e},
\label{mrc_answer_predict}
\end{aligned}
\end{equation}
where $P_i^{s}$ and $P_i^{e}$ are the model outputs defined in Eq.~\ref{mrc_prob},  $t_{\tilde{s}}$ and $t_{\tilde{e}}$ denote the start and end token of the predicted answer, respectively. By using the f1-score as the quality evaluation score of the constructed QA pairs, it is expected to generate QA pairs with high f1-score, i.e.,
\begin{equation}
    \theta^*=\mathop{\arg \max}_{\theta} \mathbb{E} f1(\mathbf{a}, \tilde{\mathbf{a}}),
\label{eq:ev_reward}
\end{equation}
where $\theta$ denotes the parameters of QS. As shown in Eq.~\ref{eq:q_pick}, the QA construction adopts the selection and concatenate operators via $\mathop{\arg \max}$ sampling technique which is discrete. 
To overcome this non-differentiable problem, we adopt policy gradient based reinforcement learning (REINFORCE)~\cite{williams1992simple} to approximate the gradients w.r.t. $\theta$. Specifically, let $\mathcal{J}(\theta)$ denote the objective function (Eq.~\ref{eq:ev_reward}), its gradient can be approximated as below:
\begin{equation}
\begin{aligned}
\nabla_{\theta}\mathcal{J}(\theta)
&\approx\frac{1}{N}\sum_{i=1}^N [\nabla_{\theta}\sum_{C_j^q\in\mathcal{N}^{\kappa}_t}\mathrm{log} P_{\theta}(C_j^q|C_t^a)  f1(\mathbf{\mathbf{a}_i}, \mathbf{f}(\mathbf{q}_i))],
\end{aligned}
\end{equation}
where $\mathbf{f}(\mathbf{q})$ transforms the question $\mathbf{q}$ into the predicted answer $\tilde{\mathbf{a}}$ via the MRC model, and $\mathcal{N}^{\kappa}_t$ denotes $\mathcal{N}^{\kappa}(C_t^a)$ for simplicity, and $P_{\theta}(C_j^q|C_t^a)$ is conditional probability of $C_j^q$ to be the corresponding question of $C_t^a$. The detail of the gradient approximation is provided in the Supplementary Material. With the approximated gradients $\nabla_{\theta}\mathcal{J}(\theta)$, the loss function for QA could be rewritten by,
\begin{equation}
\label{eq:loss_q}
\begin{aligned}
\mathcal{L}_{Q} &= -\frac{1}{N}\sum_{i=1}^N\sum_{C_j^q\in \mathcal{N}^{\kappa}_t}\mathrm{log} R_{\theta}(C_t^a, C_j^q)  [f1(\mathbf{\mathbf{a}_i}, \mathbf{f}(\mathbf{q}_i))-r_b],
\end{aligned}
\end{equation}
where $R_{\theta}(C_i^a, C_j^q) $ denotes the relevance defined in Eq.~\ref{eq:q_pick}. As only one sample per reward estimation is used, we use $r_b$ as the baseline score for reducing variance like~\cite{williams1992simple}.


\section{Experiments}
In this section, we evaluate the RMRC on three datasets by comparing it with three MRC domain adaptation methods. The code and used datasets will be released soon.

\begin{table*}[!hbt]
 \vskip -0.2cm
\renewcommand\arraystretch{1.3}
\centering
\caption{Results on d-QuAC and d-CoQA. \textbf{Bold} values indicate the best performance.}
 \vskip -0.2cm
\scalebox{0.9}{
\begin{tabular}{c|c|cc|cc|cc|cc|cc|cc} \hline
\multirow{2}{*}{\textbf{Data}} & \multirow{2}{*}{\textbf{Size}} & \multicolumn{2}{c|}{\textbf{BERT-S}}  & \multicolumn{2}{c|}{\textbf{UQACT}} & \multicolumn{2}{c|}{\textbf{CASe}} & \multicolumn{2}{c|}{\textbf{AdaMRC}} & \multicolumn{2}{c|}{\textbf{RMRC-fix}} & \multicolumn{2}{c}{\textbf{RMRC}} \\
\cline{3-14}
& & EM & F1 & EM & F1 & EM & F1 & EM & F1 & EM & F1 & EM & F1 \\
\hline
\hline
\multirow{3}{*}{d-QuAC} & ALL & 3.36 & 16.36 & 6.81 & 19.05 & 6.79 & 19.33 & 5.44 & 18.87 & 7.45 & 23.13 & \textbf{8.27} & \textbf{24.46}  \\
& 5000 & - & - & 4.26 & 17.88 & 4.85 & 17.92 & 4.57 & 18.01 & 5.85 & 21.25 & \textbf{6.37} & \textbf{22.27} \\
& 1000 & - & - & 3.62 & 16.84 & 3.91 & 17.55 & 4.12 & 17.34 & 4.42 & 19.99 & \textbf{5.67} & \textbf{21.28} \\
\hline
\multirow{3}{*}{\text{d-CoQA}} & ALL & 12.50 & 37.80 & 13.92 & 40.18 & 17.01 & 43.27 & 14.82 & 39.92 & 20.85 & 49.09 & \textbf{20.97} & \textbf{49.35} \\
& 5000 & - & - & 12.86 & 38.27 & 14.09 & 39.52 & 13.21 & 38.88 & \textbf{18.30} & \textbf{45.49} & 17.63 & 44.94 \\
& 1000 & - & - & 11.45 & 35.43 & 12.63 & 38.92 & 12.39 & 38.11 & 15.43 & 42.75 & \textbf{16.08} & \textbf{42.27} \\
\hline
\end{tabular}
 }
\label{tab:pub_dataset_result}
\vskip -0.4cm
\end{table*}

\begin{table}[!hbt]
\centering
\caption{Results on Alipay dataset.}
\vskip -0.2cm
\scalebox{0.95}{
\begin{tabular}{l|cc} 
\hline
\textbf{Model} & EM & F1 \\
\hline\hline
\textbf{BERT-S}~\cite{devlin2018bert} & 1.81 & 32.62 \\
\textbf{UQACT}~\cite{lewis2019unsupervised} & 5.32 & 42.23 \\
\textbf{CASe}~\cite{cao2020unsupervised} & 5.75 & 44.28 \\
\textbf{AdaMRC}~\cite{wang2019adversarial} & 7.20 & 43.11 \\
\hline
\textbf{RMRC-fix} & 9.25 & 52.08 \\
\textbf{RMRC} & \textbf{9.28} & \textbf{55.67} \\
\hline
\end{tabular}
}
\label{tab:alipay_dataset_result}
\vskip -0.1cm
\end{table}

\begin{table}[!hbt]
\centering
\caption{Performance of different variants of RMRC.}
\vskip -0.2cm
\begin{tabular}{l|cc} 
\hline
\textbf{Model} & EM & F1 \\
\hline\hline
BERT-S & 1.81 & 32.62 \\
RMRC w/o $r_b$  & 8.44 & 54.13 \\
RMRC w/o Answer Filtering & 2.10 & 35.05 \\
RMRC w/o Question Fusing & 8.03 & 53.71 \\
RMRC w/o QS Training  & 8.52 & 51.72 \\
RMRC w/ Confidence-based Selector & 8.76 & 52.34 \\
RMRC w/ CE Reward & 7.71 & 55.52 \\
\hline
\textbf{RMRC} & \textbf{9.28} & \textbf{55.67} \\
\hline
\end{tabular}
\label{tab:ablation_study}
 \vskip -0.4cm
\end{table}

\subsection{Datasets}
We pretrain the MRC model on the SQUAD dataset~\cite{rajpurkar2016squad} and fine-tune it on three target datasets including two public datasets (QuAC~\cite{choi2018quac} and CoQA~\cite{reddy2019coqa}) and one real-world dataset from Alipay. Note that, as the real-world Alipay data is in Chinese, we use another Chinese corpus instead of SQUAD for pre-training MRC model, i.e., a collection of CMRC~\cite{cui2018span}, DRCD~\cite{shao2018drcd} and DUREADER~\cite{he2017dureader}.

\textbf{QuAC and CoQA with Synthetic Noises: }
QuAC consists of 12,567 documents, including 11,567 training documents and 1,000 testing documents. Each document is affiliated with a related dialogue. In each round of chat in the dialogue, one user asks a question about the document, and the other user answers it. In total, there are 69,109 QA pairs for training and 5,868 for testing. 
Similar to QuAC, CoQA is composed of a training set of 7,199 documents along with 107,285 QA pairs, and a testing set of 500 documents along with 7,918 QA pairs.
As the QA pairs of QuAC and CoQA are well-matched in the dialogue, we simulate the real conversation with noisy correspondence by randomly shuffling the questions in each dialogue.
Specifically, for each dialogue, we randomly move each question ahead of the corresponding answer up to $1\sim5$ rounds. We denote the shuffled datasets of QuAC and CoQA as d-QuAC and d-CoQA, respectively in the following.

\textbf{Alipay Dataset with Real Noises: }
Alipay Dataset is collected in real-world scenarios, which contains the conversations about the marketing activities from the customer and customer service. 
In total, the dataset consists of 1,526 dialogues and 3,813 human-annotated QA pairs for testing. 

\subsection{Implementation}
In our experiments, we take the widely-used BERT as the base encoder for the QS and the MRC encoder. The BERT network contains 12 hidden layers, each of which consists of 12 attention heads. For all experiments, we generate n-grams for each document by setting $K=7$ for Eq.~\ref{eq:ngram} and set the threshold for answer filtering $\gamma$ to $0.7$ and select the question by fixing $\kappa$ in Eq.~\ref{eq:q_pick} to $5$. The optimal parameters are determined by the grid search in the Alipay dataset and used for all experiments. We set the baseline score of the reward $r_b$ to $0.7$ for Eq.~\ref{eq:loss_q} in all experiments. 
For network training, we use the Adam optimizer whose learning rate is set to $2e-5$ and $1e-5$ for pre-training and fine-tuning, respectively. More training details are provided in the supplemental material.
For evaluations, we take the \textit{Exact Match} (\textbf{EM}) and \textit{F1-score} (\textbf{F1}) as the performance measurements. 
Both the metrics are the higher the better.

\begin{figure*}[!hbt]
 \vskip -0.1cm
\centering
 \subfigure[Influence of $\gamma$] {
 \label{fig:influence_of_gamma}
 \includegraphics[width=0.31\textwidth]{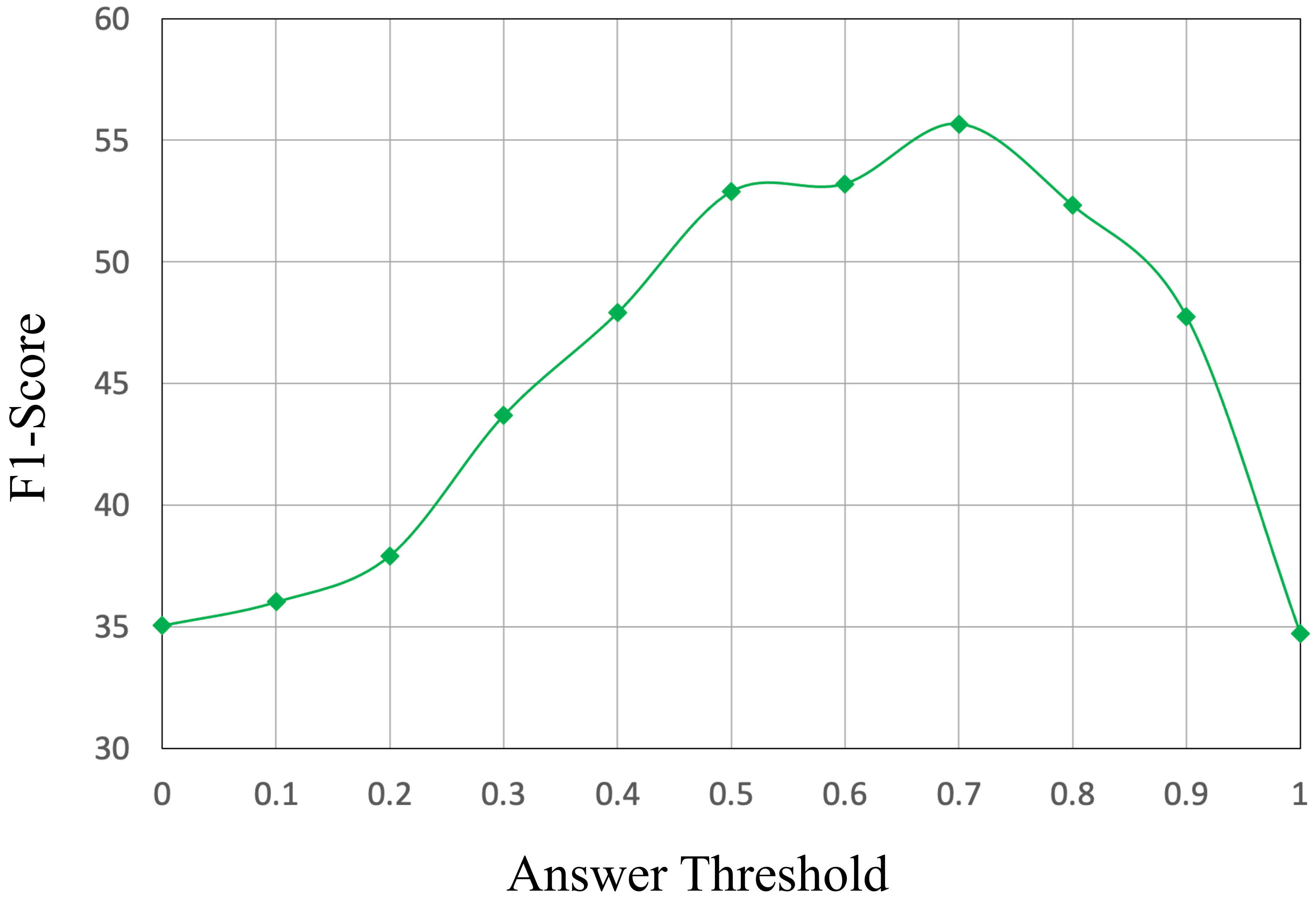}
 }
 \subfigure[Influence of $\kappa$] {
 \label{fig:influence_of_topk}
 \includegraphics[width=0.31\textwidth]{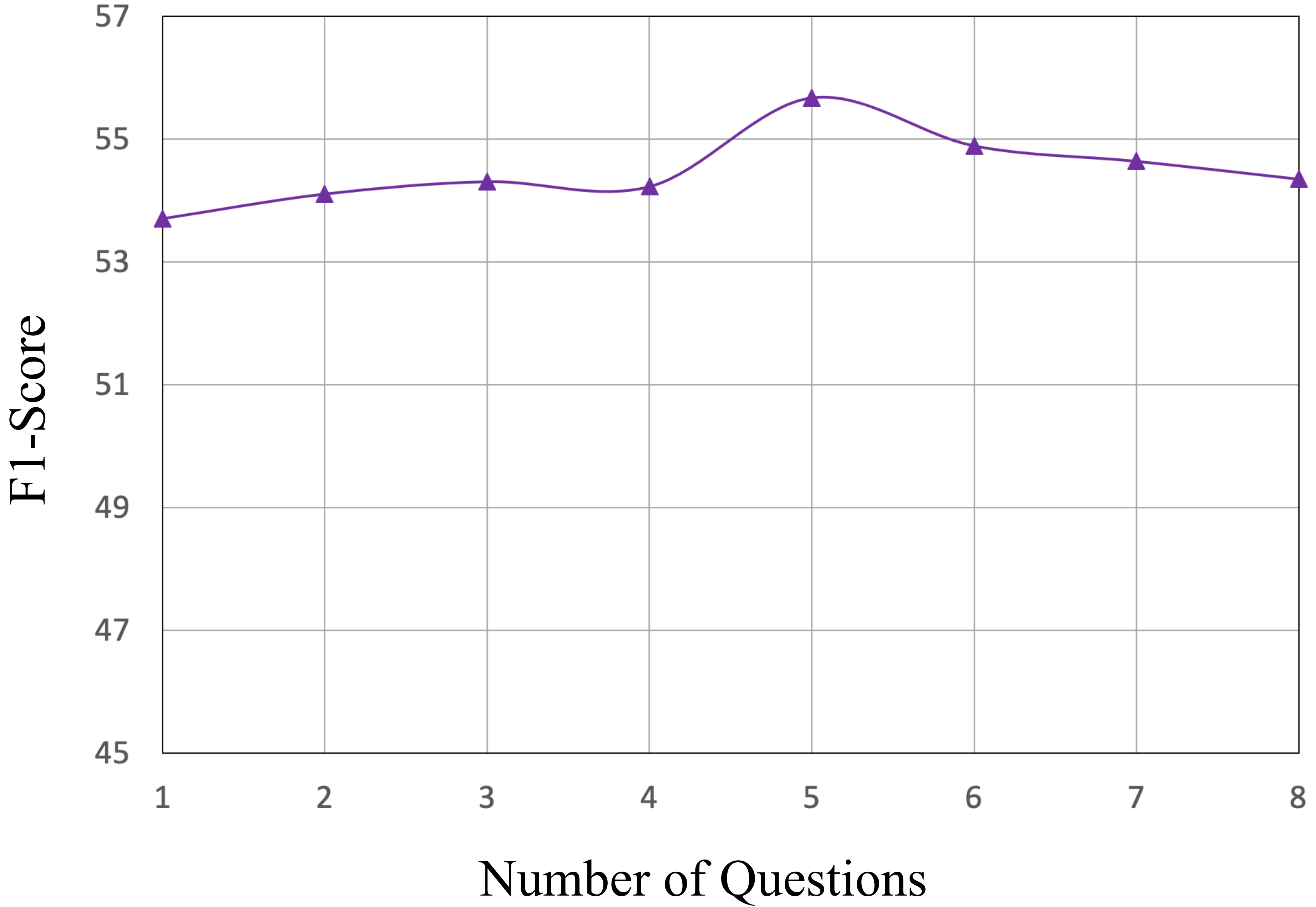}
 }
 \subfigure[Influence of $r_b$] {
 \label{fig:influence_of_baseline}
 \includegraphics[width=0.31\textwidth]{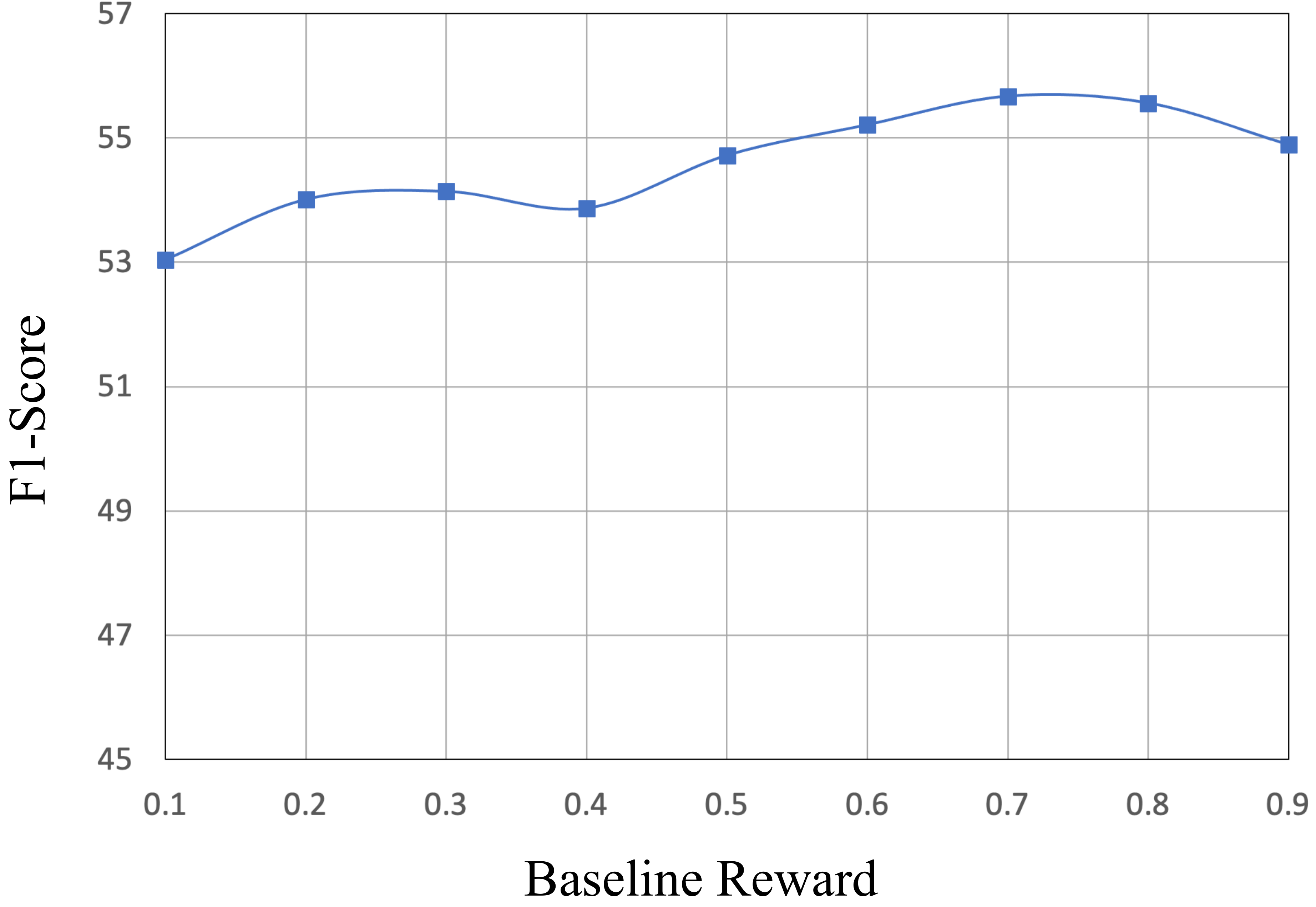}
 }
 \vskip -0.3cm
\caption{(a-c) Influence of answer filtering threshold ($\gamma$), the selected question number ($\kappa$), and the baseline reward ($r_b$). }\label{fig:paras}
\vskip -0.2cm
\end{figure*}

\begin{figure*}[!hbt]
\centering
\includegraphics[width=1.0\textwidth]{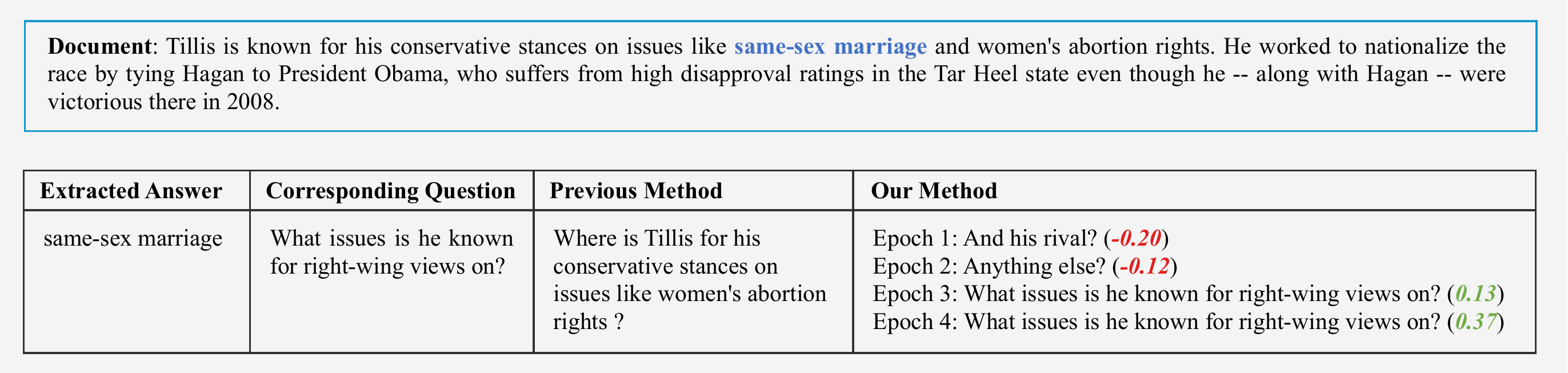}
\caption{Comparisons between the questions constructed by our method and existing work~\cite{lewis2019unsupervised}. The \textcolor{red}{red number} and \textcolor{green}{green number} denote the negative and positive reward from MRC model, respectively.}
\label{fig:case_study}
 \vskip -0.3cm
\end{figure*}

\subsection{Comparison Experiments}
To show the effectiveness of our method, we compare it with the baselines including a vanilla BERT model trained in the source domain (denoted by BERT-S) and three QA generation based MRC domain adaptation methods, i.e., UQACT~\cite{lewis2019unsupervised}, AdaMRC~\cite{wang2019adversarial} and CASe~\cite{cao2020unsupervised}. Note that CASe~\cite{cao2020unsupervised} uses the annotated questions in the target domain for fine-tuning, while the questions are unavailable in our experiment settings. As a remedy, we use a pre-trained question generator to generate questions in the target domain for CASe. All the baselines and our method follow the same training pipeline, i.e., first pretraining the MRC model on the SQUAD data, then fine-tuning and evaluating it on the target datasets (d-QuAC, d-CoQA and Alipay). For all baselines, we conduct the experiments with the recommended parameters and report the best one. For our method, we repeat the experiments three times with different seeds and report the average. 
Moreover, we additionally use a variant of RMRC with a fixed QS (denoted as RMRC-fix) that keep the weights of QS fixed instead of optimizing with proposed reinforce self-training optimizer. Such a baseline could help us to understand the influence of the proposed reinforce self-training strategy.

\textbf{Results on d-QuAC and d-CoQA}: Table~\ref{tab:pub_dataset_result} shows the quantitative results on d-QuAC and d-CoQA.
From the results, one could observe that all domain adaptation methods outperform BERT-S. 
Moreover, the proposed methods (i.e., RMRC and RMRC-fix) significantly outperform the other three domain adaptation methods.  Especially, RMRC achieves performance gains of \textbf{1.48/5.13} and \textbf{3.96/6.08} in d-QuAC-ALL and d-CoQA-ALL compared with the best baseline, respectively. From the comparisons between RMRC and RMRC-fix, one could easily find the effectiveness of our QS training strategy.
Furthermore, to investigate the performance of the proposed method with small training datasets, we conduct experiments on QuAC and CoQA with $5,000$ and $1,000$ training samples. In other words, only 1.45\% of original training QuAC samples are used in the evaluation with d-QuAC-1000. As shown, with the data size decreasing, the performance of all methods decreases while RMRC still outperforms all baselines.

\textbf{Results on Alipay Dataset}: Table~\ref{tab:alipay_dataset_result} shows the quantitative results on the Alipay dataset. From the results, one could find that RMRC outperforms all baselines by a considerable performance margin, demonstrating its effectiveness in real-world dataset. Specifically, RMRC achieves performance gains of \textbf{2.08} and \textbf{12.56} over the best baseline (CASe) in terms of EM and F1, respectively.

\subsection{Ablation Study}
In this section, we carry out ablation study on the Alipay dataset. In the experiments, we report the performance by removing or replacing some modules, including removing answer filtering (RMRC w/o Answer Filtering), question fusing (RMRC w/o Question Fusing), QS training (RMRC w/o QS Training); replacing f1-score reward by cross-entropy reward (RMRC w/ CE Reward), and reinforced self-training mechanism by confidence-based self-training mechanism (RMRC w/ Confidence-based Selector). 
As shown in Table~\ref{tab:ablation_study}, all the proposed modules are crucial to achieving encouraging results and the following conclusions could be obtained: i) All the RMRC variants outperform BERT-S, demonstrating the value of the dialogue-based QA pair construction; ii) RMRC and RMRC w/ CE Reward significantly outperform both RMRC w/o QS Training and RMRC w/ Confidence-based Selector, which shows the effectiveness of the proposed reinforced self-training algorithm for QS training; iii) RMRC that using F1 as the reward achieves better performance CE as F1 score directly measures the precision of the predicted answer.

\subsection{Parameter Sensitivity Analysis}
In this section, we carry out experiments on the Alipay dataset to investigate the influence of different parameters in RMRC including the answer filtering threshold $\gamma$ in Eq.~\ref{eq:a_pick}, the number of nearest questions $\kappa$ in Eq.~\ref{eq:q_pick}, and the baseline reward $r_b$ in Eq.~\ref{eq:loss_q}. 
As shown in Fig.~\ref{fig:influence_of_gamma}, the performance keeps continuously increasing with increasing $\gamma$ and gets the best result when $\gamma=0.7$. As shown in Fig.~\ref{fig:influence_of_topk}, the performance of the MRC model keeps continuously increasing with increasing $\kappa$ from $1$ to $5$, showing that the correct question was more likely selected with more selected questions. As $\kappa$ increases from $5$ to $7$, the performance significantly decreases. The reason is that too many selected questions inevitably introduce more noisy pairs. As shown in Fig.~\ref{fig:influence_of_baseline}, the performance of the MRC model keeps improving when $r_b$ increases from $0.1$ to  $0.7$, and then decreases when $r_b$ is larger than $0.7$. The reason is that an over-high $r_b$ will discourage almost all the predictions for QS training, thus leading to incorrect feedback for QS. 

\subsection{Case Study}
To investigate the effectiveness of the QS and reinforced self-training optimizer, we conduct the case study in Fig.~\ref{fig:case_study}. As shown, the previous domain adaptation works will suffer from the NC problem, thus yielding irrelevant questions to the answer. 
In contrast, RMRC explicitly solves the NC problem by training QS with the MRC model feedback. Specifically, in the first epoch, QS would select an irrelevant question with a negative reward from the MRC model. After training with several epochs, the optimized QS could find the desirable question to the answer. Such an example intuitively demonstrates the effectiveness of our QS and the reinforced self-training algorithm.

\section{Conclusion}
This paper could be the first successful attempt to solve the noisy correspondence problem in the domain adaptation methods for MRC. Different from the well-studied noisy labels, the noisy correspondence refers to the errors in alignment instead of category-level annotation. To overcome this challenge, we propose a robust domain adaptation method for MRC with a novel reinforced self-training optimizer. Extensive experiments verify the effectiveness of the proposed method in leveraging synthesized and real-world dialogue data for MRC. 


\bibliography{ref} 

%

\end{document}